\documentclass[iicol,pdflatex,sn-basic,Numbered]{sn-jnl}

\usepackage{graphicx}%
\usepackage{multirow}%
\usepackage{amsmath,amssymb,amsfonts}%
\usepackage{amsthm}%
\usepackage{mathrsfs}%
\usepackage[title]{appendix}%
\usepackage{xcolor}%
\usepackage{textcomp}%
\usepackage{manyfoot}%
\usepackage{booktabs}%
\usepackage{algorithm}%
\usepackage{algorithmicx}%
\usepackage{algpseudocode}%
\usepackage{listings}%

\theoremstyle{thmstyleone}%
\theoremstyle{thmstyletwo}%

\theoremstyle{thmstylethree}%

\begin{document}

\title[mocorp]{MoCoRP: Modeling Consistent Relations between Persona and Response for Persona-based Dialogue}

\author[1]{\fnm{Kyungro} \sur{Lee}}\email{lkr981147@gm.gist.ac.kr}
\author[1]{\fnm{Dongha} \sur{Choi}}\email{dongha528@gm.gist.ac.kr}
\author*[1]{\fnm{Hyunju} \sur{Lee}}\email{hyunjulee@gist.ac.kr}

\affil[1]{\orgdiv{Artificial Intelligence Graduate School}, \orgname{Gwangju Institute of Science and Technology}, \orgaddress{\city{Gwangju}, \postcode{61005}, \country{South Korea}}}

\abstract{
As dialogue systems become increasingly important across various domains, a key challenge in persona-based dialogue is generating engaging and context-specific interactions while ensuring the model acts with a coherent personality. However, existing persona-based dialogue datasets lack explicit relations between persona sentences and responses, which makes it difficult for models to effectively capture persona information. To address these issues, we propose \textbf{MoCoRP} (\textbf{Mo}deling \textbf{Co}nsistent \textbf{R}elations between \textbf{P}ersona and Response), a framework that incorporates explicit relations into language models.
MoCoRP leverages an NLI expert to explicitly extract the NLI relations between persona sentences and responses, enabling the model to effectively incorporate appropriate persona information from the context into its responses. We applied this framework to pre-trained models like BART and further extended it to modern large language models (LLMs) through alignment tuning.
Experimental results on the public datasets ConvAI2 and MPChat demonstrate that MoCoRP outperforms existing baselines, achieving superior persona consistency and engaging, context-aware dialogue generation. Furthermore, our model not only excels in quantitative metrics but also shows significant improvements in qualitative aspects. These results highlight the effectiveness of explicitly modeling persona-response relations in persona-based dialogue. The source codes of MoCoRP are available at \url{https://github.com/DMCB-GIST/MoCoRP}.
}


\maketitle

\bmhead{Acknowledgements}
We appreciate the high-performance GPU computing support provided by HPC-AI Open Infrastructure via GIST SCENT, as well as by the Artificial Intelligence Industry Cluster Agency (AICA).

\section{Introduction} \label{intro}
With recent advances in natural language processing (NLP)~\cite{devlin-etal-2019-bert, brown2020languagemodelsfewshotlearners}, there has been an increasing focus on developing human-like dialogue systems.
Researchers have explored incorporating a persona, typically represented as a textual description, into models to create personalized chatbots~\cite{li-etal-2016-persona, ma-etal-2021-one, zhong-etal-2022-less}.
In persona-based dialogue, the model is required to generate a response grounded in both the dialogue history and persona sentences, while ensuring persona consistency and continuing the conversation in an engaging and context-specific manner~\cite{wolf2019transfertransfotransferlearningapproach, song-etal-2021-bob}.

\begin{figure}[t!]
    \centering
    \includegraphics[width=\linewidth]{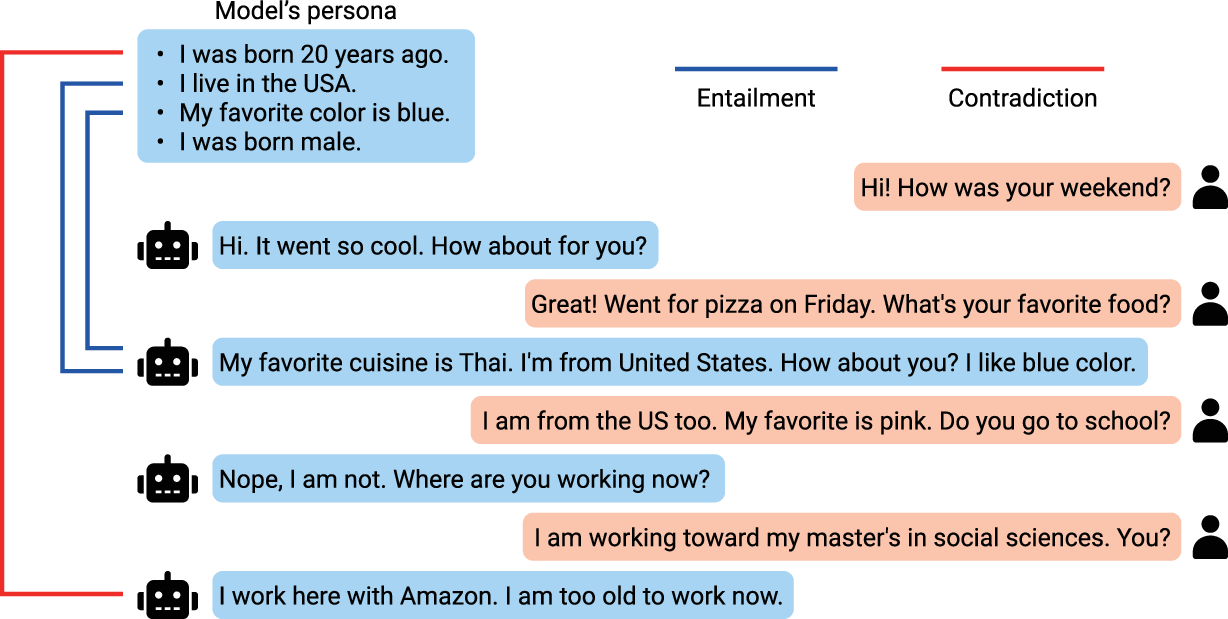}
    \caption{Example of persona-based dialogue from the ConvAI2 dataset.
    The relations between the model's persona and utterances are represented by blue lines for entailment and red lines for contradiction, while relations not indicated are neutral.
    In the original dataset, the relations between persona sentences and response are not provided.}
    \label{fig:convai}
\end{figure}

A widely known persona-based dialogue dataset is ConvAI2~\cite{dinan-etal-2020-convai, zhang-etal-2018-personalizing}, in which two annotators take on provided persona sentences and engage in conversations to get to know each other. However, these datasets lack explicit guidance on how each persona sentence relates to the corresponding response. Our analysis of these datasets using a natural language inference (NLI) model confirms that the majority of persona-response pairs are neutral relations, while cases of entailment are infrequent and contradictions are highly uncommon (see Figure~\ref{fig:convai}). This absence of explicit relations and the prevalence of neutral cases make it difficult for dialogue models to effectively capture persona information and maintain consistency.

To address this, we hypothesize that dialogue agents need the ability to recognize the explicit relationships between persona sentences and target responses. We propose \textbf{Mo}deling \textbf{Co}nsistent \textbf{R}elations between \textbf{P}ersona and Response for Persona-based Dialogue (\textbf{MoCoRP}), a framework that leverages an NLI expert's relation prediction capability to produce explicit NLI relations. This allows the model to refer these relations when generating responses, thus improving its ability to maintain persona consistency. We initially apply our approach to a language model like BART~\cite{lewis-etal-2020-bart} and, through alignment tuning, confirm its applicability to modern large language models (LLMs). Our model achieves superior performance on quantitative metrics, while also showing significant improvements in qualitative aspects.

Our key contributions are as follows:
\begin{itemize}
    \item We analyzed the NLI relations between persona sentences and target responses in persona-based dialogue datasets, finding that neutral relations constitute the majority of cases.
    \item We propose MoCoRP, a novel framework that incorporates an NLI expert's relation prediction capability into language models and confirms its applicability to large language models via alignment tuning.
    \item We conduct comparative experiments on ConvAI2 and MPChat~\cite{ahn-etal-2023-mpchat}, demonstrating that our approach surpasses baselines and achieves superior performance on both quantitative and qualitative metrics.
\end{itemize}

\section{Related Works} \label{rel_work}

\subsection{Persona-based Dialogues}
Various methodologies have been proposed to build the persona-based dialogue models.
Early studies adopted an approach of forming speaker embeddings and incorporating them into the model alongside word embeddings to generate personalized responses~\cite{li-etal-2016-persona}.
Mazar{\'e} et al.~\cite{mazare-etal-2018-training} constructed a large-scale persona-based dialogue dataset extracted from Reddit and used it for additional fine-tuning.
With the advent of large-scale pre-trained language models, research on fine-tuning such models has significantly increased.
Wolf et al.~\cite{wolf2019transfertransfotransferlearningapproach} and Golovanov et al.~\cite{golovanov-etal-2019-large} demonstrated strong performance in the ConvAI2 competition using GPT-based models.
Liu et al.~\cite{liu-etal-2020-impress} improved persona consistency through mutual persona perception, focusing on understanding between the interlocutors.
Zheng et al.~\cite{zheng2020pre} proposed an attention routing mechanism in transformer-based models to assign weights to persona and context information.
Additionally, there have been attempts to enhance persona-based dialogue generation performance through text manipulation and data augmentation using pre-trained models~\cite{cao-etal-2022-model}.

\subsection{Natural Language Inference}
Efforts have been made to address consistency issues in dialogue systems by leveraging natural language inference (NLI)~\cite{bowman-etal-2015-large, williams-etal-2018-broad, welleck-etal-2019-dialogue}.
This task involves determining whether a hypothesis can be logically derived from a given premise as entailment, neutral, or contradiction.
The Dialogue NLI dataset~\cite{welleck-etal-2019-dialogue} was constructed from the PersonaChat dataset~\cite{zhang-etal-2018-personalizing} to improve consistency in dialogue models, while the DECODE dataset~\cite{nie-etal-2021-like} was specifically developed for contradiction detection.
Research using such datasets has focused on improving persona consistency in dialogue agents.
BoB~\cite{song-etal-2021-bob} improved persona consistency understanding through unlikelihood training using a non-dialogue NLI dataset.
LMEDR~\cite{chen-etal-2023-learning} employed two independent external memory structures learned from both NLI and dialogue datasets to effectively incorporate entailment and discourse relations.
Zhu et al.~\cite{zhu2023personalized} proposed a classifier capable of selecting relevant persona sentences from the context and refined responses through persona selection and consistency inference.

\begin{figure*}[t!]
    \centering
    \includegraphics[width=\textwidth]{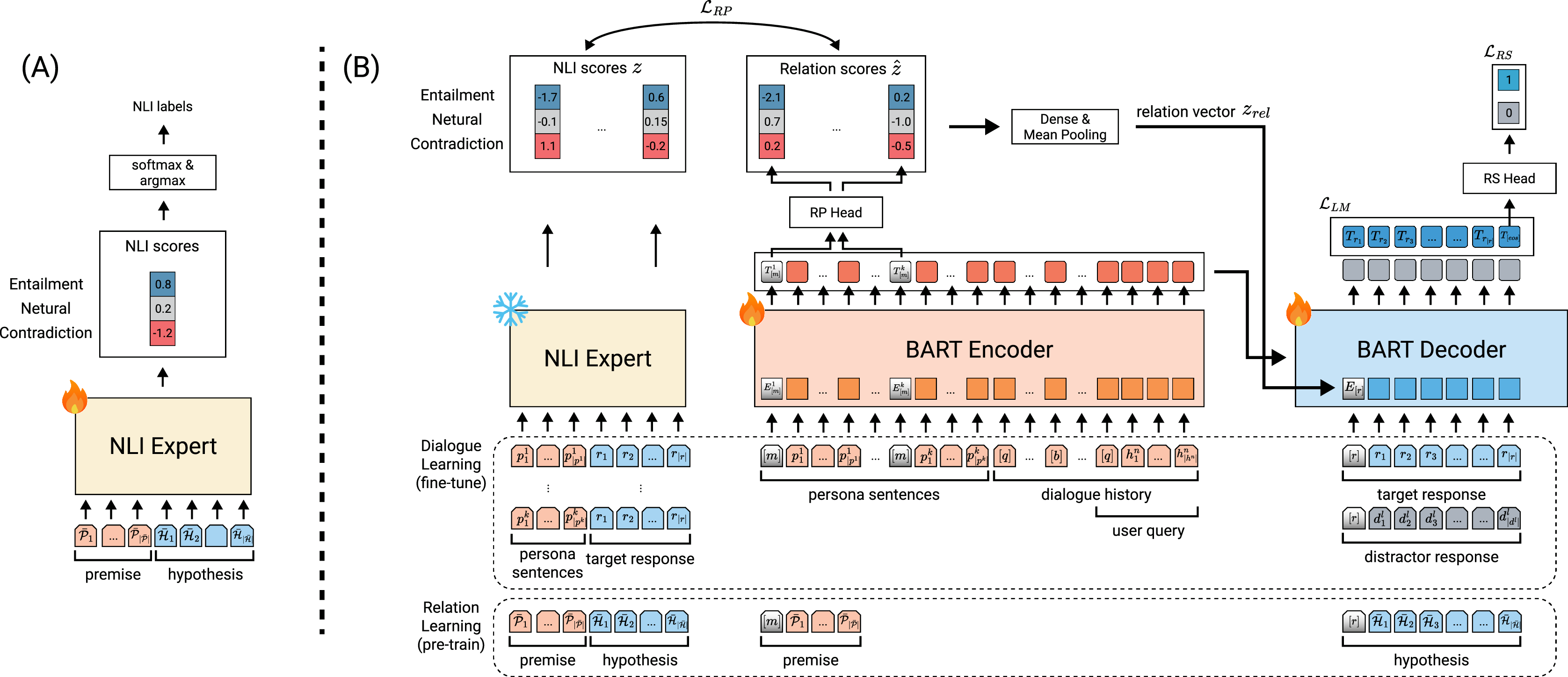}
    \caption{Overall architecture of the proposed MoCoRP for persona-based dialogue.
    The NLI expert is trained to predict NLI labels using the Dialogue NLI dataset (A), and BART learns the relation prediction capability from the NLI expert through relation learning and dialogue learning (B).
    Special tokens are used to structure the input: $[m]$ represents the mask token from the tokenizer, $[q]$ indicates that the following token sequence is a user query, and $[b]$ marks a previous bot utterance.
    The BART decoder start token, $[r]$, is transformed into a token embedding through the embedding layer.
    This embedding is then combined with the relation vector $z_{rel}$ to model the relations between the sentences associated with the $[m]$ and the decoder input.}
    \label{fig:mocorp}
\end{figure*}

\section{Method} \label{met}
In this section, we introduce the MoCoRP, which explicitly models the relations between persona sentences and target response.
This section begins with a definition of the task (Section~\ref{met:task_def}) and then describes the overall pipeline of MoCoRP (Section~\ref{met:mocorp}).
Specifically, Section~\ref{met:mocorp:nli_exp} discusses training and utilizing an NLI expert to capture the NLI relations between persona sentences and the target response.
Section~\ref{met:mocorp:rel_learn} explains the first stage task, relation learning, which uses an NLI dataset.
The second stage task, dialogue learning, involving fine-tuning with a persona-based dialogue dataset, is presented in Section~\ref{met:mocorp:diag_learn}.
We then discuss the training objective for MoCoRP in Section~\ref{met:mocorp:tra_obj}.
Additionally, we introduce a method to extend this framework to large language models (MoCoRP LLM, Section~\ref{met:mocorpllm}).

\subsection{Task Definition} \label{met:task_def}
Formally, the dialogue history can be defined as $H=\{h^1, h^2, \dots, h^m, \dots, h^n\}$, where $h^m$ is the $m$-th utterance of the conversation representing either the user query or the previous bot utterance.
Note that the dialogue history ends with the user query $h^n$, and the model is expected to provide a response that aligns with this query.
In the current dialogue, there are $k$ persona sentences denoted as $P=\{p^1, p^2, \dots, p^k\}$, the gold response $R=\{r\}$, and a set of $l$ randomly selected distractor responses $D=\{d^1, d^2, \dots, d^l\}$.
Additionally, we use premise-hypothesis sentence pairs from the NLI dataset $\mathcal{N}=\{\bar{\mathcal{P}},\bar{\mathcal{H}}\}$.
The relation between the two sentences is categorized as entailment, neutral, or contradiction, depending on whether the hypothesis can be inferred from the premise, is unrelated, or contradicts it, respectively.

The task of persona-based dialogue is defined by language modeling and response selection.
Language modeling focuses on generating a persona-consistent response $\hat{R}=\{\hat{r} \}$ based on the given persona sentences $P$ and dialogue history $H$.
Response selection, similar to next sentence prediction~\cite{devlin-etal-2019-bert}, involves selecting the correct next dialogue utterance between the gold response $R$ and $l$ distractor responses $D$.

\subsection{MoCoRP} \label{met:mocorp}
Figure~\ref{fig:mocorp} illustrates the overall architecture of MoCoRP, which employs an encoder-decoder transformer~\cite{NIPS2017_transformer} based on BART~\cite{lewis-etal-2020-bart}.
The MoCoRP is trained through an iterative two-stage approach.
In each epoch, it is first pre-trained on an NLI dataset and then fine-tuned on a persona-based dialogue dataset.

During training, MoCoRP explicitly models the relations between persona sentences and the target response.
To enable the BART model to learn the NLI expert’s relation prediction capability, we introduce a relation prediction task.
This task aligns the relation scores $\hat{z}$, computed by the BART encoder, with the NLI scores $z$, computed by the NLI expert, to represent the relations between persona sentences and the target response.
With explicit relation information, the model can selectively incorporate information from multiple persona sentences into the target response.
Once the training of MoCoRP is complete, the model is expected to generate persona-consistent responses even in the absence of explicit relation information.

\subsubsection{Training NLI Expert} \label{met:mocorp:nli_exp}
To obtain relations between persona sentences and target response that do not present existing persona-based dialogue dataset, we leverage NLI.
Specifically, we introduce the NLI expert, a sentence classifier based on RoBERTa~\cite{liu2019robertarobustlyoptimizedbert}.
As illustrated in Figure~\ref{fig:mocorp} (A), the NLI expert takes two sentences as input and classifies their relation as entailment, neutral, or contradiction.
It is trained using premise-hypothesis pairs from the Dialogue NLI dataset.

We utilize the NLI expert in two ways. 
First, it is employed to analyze persona-based dialogue datasets. 
Second, it is used to estimate the "NLI score," which serves as a guide for the relation prediction task described in the following subsections.

\subsubsection{Relation Learning} \label{met:mocorp:rel_learn}
The NLI expert was used to analyze the relations between persona sentences and target response in persona-based dialogue datasets.
The results show that neutral is the most common relation, accounting for 84\% in ConvAI2 and 72\% in MPChat (see Appendix~\ref{app:stat_rel}).
This imbalance may lead the model to predominantly learn neutral relations, limiting its ability to effectively use relation information.
To mitigate this bias, we introduce a pre-training task, relation learning, which samples a subset of the NLI dataset.
By pre-training balanced relations through relation learning before training in a dialogue setting, the model can better represent a diverse range of NLI relations.

In the given NLI dataset, the BART encoder takes the premise $\bar{\mathcal{P}}$ as input, while the decoder learns the vector representation of the hypothesis $\bar{\mathcal{H}}$.
Formally, the inputs to the encoder and decoder follow $\big[ [m], \bar{\mathcal{P}}_{1}, \bar{\mathcal{P}}_{2}, \dots, \bar{\mathcal{P}}_{|\bar{\mathcal{P}}|} \big]$ and $\big[ [r], \bar{\mathcal{H}}_{1}, \bar{\mathcal{H}}_{2}, \dots, \bar{\mathcal{H}}_{|\bar{\mathcal{H}}|} \big]$, where $[m]$ and $[r]$ represent the mask token of tokenizer and decoder start token, respectively, indicating the start of the premise and hypothesis.

To learn the NLI relation between the premise and hypothesis, the hidden representation of the mask token, $T_{[m]}^1$, is obtained by the BART encoder.
This representation is then passed to the relation prediction head (RP Head) to obtain a relation score $\hat{z}^{RL}$,
\begin{gather}
    \hat{z}^{RL} = \text{RP Head}(T_{[m]}^1) \in \mathbb{R}^3 
    \label{eq1}
\end{gather}
where the RP Head is a down-projection fully connected layer, mapping the hidden representation into a 3-dimensional space representing entailment, neutral, and contradiction.

To enable BART to inherit the relation prediction capability of the NLI expert, we leverage the NLI expert in the training process.
The NLI expert generates NLI score $z^{RL}\in\mathbb{R}^3$, which serves as a training signal for learning the NLI relation between a premise and a hypothesis.
The NLI scores $z^{RL}$ are obtained by providing the NLI expert with premise $\bar{\mathcal{P}}$ and hypothesis $\bar{\mathcal{H}}$ pairs from the training set of the NLI dataset into the NLI expert.
By aligning the relation score $\hat{z}^{RL}$ with the NLI score $z^{RL}$, BART can effectively capture the relation between sentences.

The relation score $\hat{z}^{RL}$ is transformed into a relation vector $z_{rel}^{RL}$ via a dense layer and added to the decoder start token embedding $E_{[r]}$, as proposed in Chen et al.~\cite{chen-etal-2023-learning}:
\begin{gather}
    z_{rel}^{RL} = \text{Dense} (\hat{z}^{RL}) \in \mathbb{R}^{d_{model}} \\
    \bar{E}_{[r]} = E_{[r]} + z_{rel}^{RL}
    \label{eq2:3}
\end{gather}
where the Dense layer is an up-projection fully connected layer, transforming the relation score to match the hidden dimension of the model $d_{model}$. The model then proceeds from $\bar{E}_{[r]}$ through each decoder transformer layer to produce a final representation.

The model is trained using multiple loss functions.
First, language modeling loss involves generating the hypothesis $\bar{\mathcal{H}}$ from the given premise $\bar{\mathcal{P}}$.
Second, response selection loss determines whether the relation between the two sentences is entailment or not.
This can facilitate selecting the next dialogue utterance during dialogue learning.
Lastly, relation prediction loss aligns the relation score vector $\hat{z}^{RL}$ with the NLI score vector $z^{RL}$.
Further details of the training objective are discussed in Section~\ref{met:mocorp:tra_obj}.

\subsubsection{Dialogue Learning} \label{met:mocorp:diag_learn}
Once relation learning is complete, BART is fine-tuned on the persona-based dialogue dataset.
The BART encoder takes the persona sentences $P$ and dialogue history $H$ as input, while the decoder learns the vector representation of the response $R$.
Formally, the input representation of the encoder follows $\big[ [m], p^1, [m], p^2, \dots, [m], p^k, [q], h^1, [b], h^2, \dots, [q], h^n \big]$, where the mask token $[m]$ is prepended to each persona sentence to indicate it.

Similar to relation learning, from the last hidden states $T$ computed by the encoder, the $k$ vectors corresponding to each mask token, $[T^1_{[m]}, T^2_{[m]}, \dots, T^k_{[m]}]$, are passed through the RP Head to compute relation scores $\hat{z}$.
Subsequently, the $k$ relation scores are processed through a dense layer and mean pooling to obtain the relation vector $z_{rel}$, which is then added to the embedding vector of the decoder start token $E_{[r]}$:
\begin{gather}
    \hat{z}^i = \text{RP Head}(T_{[m]}^i) \in \mathbb{R}^3 \\
    z_{rel} = \frac{1}{k} \sum_{i=1}^k \text{Dense} (\hat{z}^i) \in \mathbb{R}^{d_{model}} \\
    \bar{E}_{[r]} = E_{[r]} + z_{rel}.
    \label{eq4:6}
\end{gather}
Here, the decoder starts from the embedding vector $\bar{E}_{[r]}$ to produce the final representation.

In a manner similar to relation learning, BART captures the relations between the persona sentences and target response during dialogue learning.
This is achieved by aligning the relation scores $\hat{z}$ with the NLI scores $z$ generated by the NLI expert.
The NLI expert takes pairs of $k$ persona sentences $P$ and target response $R$ from the training set of the persona-based dialogue dataset, producing $k$ NLI scores $z\in\mathbb{R}^{k\times3}$.

The model is optimized using three types of loss: language modeling, response selection, and relation prediction.
The language modeling loss focuses on predicting the target response $R$ based on persona sentences $P$ and dialogue history $H$.
The response selection loss involves selecting the next dialogue utterance from the gold response $R$ and distractor responses $D$ based on the persona sentences and dialogue history.
Finally, the relation prediction loss aligns the relation score vectors $\hat{z}$ with the NLI score vectors $z$.

\subsubsection{Training Objective} \label{met:mocorp:tra_obj}
The model is trained using the negative log-likelihood for the language modeling loss.
In relation learning, as shown in Equation~\ref{eq7}, the language modeling loss incorporates the relation vector $z_{rel}^{RL}$ to generate the hypothesis $\bar{\mathcal{H}}$ from the premise $\bar{\mathcal{P}}$.
Similarly, during dialogue learning, it involves generating the target response $R$ from the given persona sentences $P$, dialogue history $H$, and the relation vector $z^{rel}$, as shown in Equation~\ref{eq8}:
\begin{equation}
    \begin{aligned}
        \mathcal{L}_{LM}^{RL} &= -\log{ p_{\theta} (\bar{\mathcal{H}} | \bar{\mathcal{P}}, z_{rel}^{RL})} \\
        &= -\sum_{i=1}^{|\bar{\mathcal{H}|}} \log{ p_{\theta} ( \bar{\mathcal{H}}_i | \bar{\mathcal{P}}, z_{rel}^{RL}, \bar{\mathcal{H}}_{<i}) }
    \end{aligned}
    \label{eq7}
\end{equation}
\begin{equation}
    \begin{aligned}
        \mathcal{L}_{LM} &= -\log{ p_{\theta} (R|P, H, z_{rel})} \\
        &= -\sum_{i=1}^{|R|} \log{ p_{\theta} (R_i | P, H, z_{rel}, R_{<i}) }.
    \end{aligned}
    \label{eq8}
\end{equation}

For the response selection loss, we use the value $\hat{y}$ obtained by processing the eos token hidden representation $T_{[eos]}$ from the BART Decoder.
This is done through the response selection head (RS Head), following a similar approach to Wolf et al.~\cite{wolf2019transfertransfotransferlearningapproach}.
In relation learning, selecting an entailment hypothesis for a given premise results in the label $y$ being positive, while selecting a neutral or contradiction hypothesis results in the label $y$ being negative.
Similarly, in dialogue learning, selecting the gold response based on the given persona sentences and dialogue history assigns a positive label to $y$, whereas selecting a distractor response assigns a negative label.
This is optimized using binary cross entropy loss:
\begin{gather}
    \hat{y} = \text{RS Head}(T_{[eos]}) \\
    \mathcal{L}_{RS} = - [y \log(\hat{y}) + (1 - y) \log(1 - \hat{y})].
    \label{eq9:10}
\end{gather}

To explicitly model the NLI relations, the relation prediction loss is introduced, aiming to minimize the KL divergence between sentence pairs.
In relation learning, the model aligns the relation score vector $\hat{z}^{RL}$, produced by the BART encoder, with the NLI score vector $z^{RL}$, computed by the NLI expert.
In case of dialogue learning, the relation score vectors $\hat{z}$ and the NLI score vectors $z$ are from the persona sentences and target response.
Since both BART and the NLI expert process identical sentence pairs in each task—premise and hypothesis in relation learning, and persona sentences and target response in dialogue learning—aligning the relation score vector $\hat{z}$ closely with the NLI score vector $z$ enables BART to acquire the relation prediction capability of the NLI expert.
This alignment allows the model to leverage necessary persona information from context during inference without explicit relations.
\begin{gather}
    P(z) = \text{softmax}(z),\quad Q(\hat{z}) = \text{softmax}(\hat{z}) \\
    \mathcal{L}_{RP} = D_{\text{KL}}\big( P(z) \parallel Q(\hat{z}) \big) = \sum_{i=1}^3 P_i(z)\log{\frac{P_i(z)}{Q_i(\hat{z})}}
    \label{eq11:12}
\end{gather}

Relation learning and dialogue learning are each optimized by minimizing the following objectives:
\begin{gather}
    \mathcal{L}^{RL} = \mathcal{L}_{LM}^{RL} + \mathcal{L}_{RS}^{RL} + \alpha \mathcal{L}_{RP}^{RL} \\
    \mathcal{L} = \mathcal{L}_{LM} + \mathcal{L}_{RS} + \alpha \mathcal{L}_{RP}
    \label{eq13:14}
\end{gather}
where $\alpha$ is a hyperparameter chosen from the range [0, 1].

Finally, MoCoRP is pre-trained through relation learning and subsequently fine-tuned through dialogue learning on a persona-based dialogue dataset.

\begin{figure*}[t!]
    \centering
    \includegraphics[width=\textwidth]{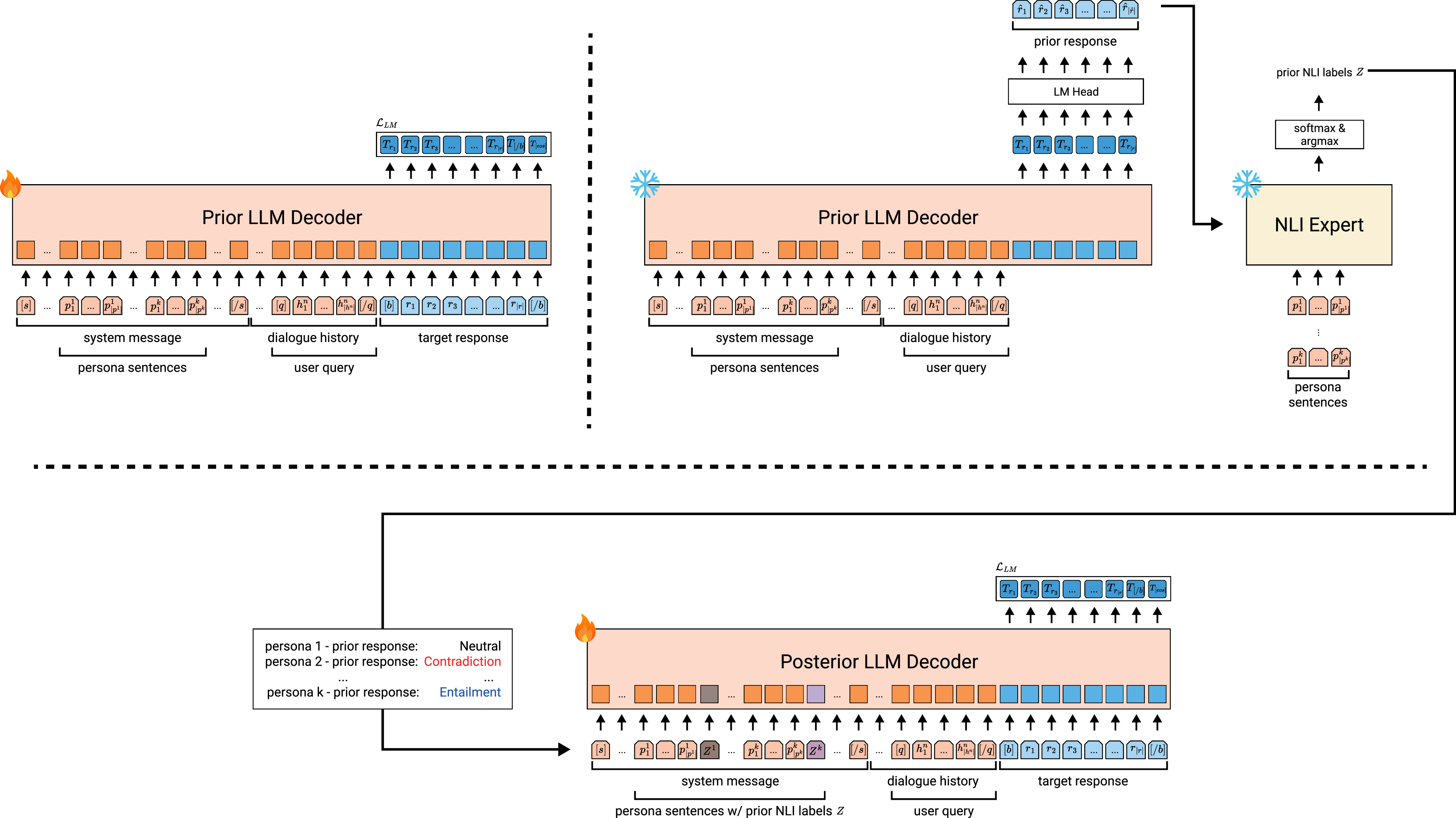}
    \caption{Overall architecture of the proposed MoCoRP LLM for persona-based dialogue.
    The prior LLM is trained to generate the target response based on the given system message and dialogue history (top left).
    After completing alignment tuning of the prior model, the NLI expert calculates the NLI labels between the persona sentences and the prior response generated by the prior LLM (top right).
    Using these NLI labels along with the given input, the posterior model learns to generate the target response by maximizing its probability conditioned on the input context (bottom).}
    \label{fig:mocorpllm}
\end{figure*}
\subsection{MoCoRP LLM} \label{met:mocorpllm}
Traditionally, the area of persona-based dialogue research has been limited to standard language models, which lack scalability.
To address this, we extend the MoCoRP framework to large language models (LLM), leveraging alignment tuning~\cite{NEURIPS2022_rlhf}.
Unlike conventional language models, LLMs are pre-trained on massive corpora~\cite{kaplan2020scalinglawsneurallanguage, hoffmann2022trainingcomputeoptimallargelanguage}.
This extensive training aligns the model architecture and optimization process closely with the scale of the data.
Therefore, if the model architecture is altered and is not trained with sufficiently large data to match the changes, unexpected performance drops may occur due to the imbalance between the architecture and data scale.
Therefore, this section describes methods for utilizing explicit relations between persona sentences and target response without modifying the LLM architecture.

As shown in Figure~\ref{fig:mocorpllm}, MoCoRP LLM consists of a prior and a posterior model, leveraging the NLI expert as in MoCoRP.
With their powerful language generation capabilities, LLMs can generate responses that partially reflect personas even in the absence of explicit NLI relations.
The proposed MoCoRP LLM extracts textual NLI relations $Z$ between the prior response, generated by the prior model without NLI relations, and persona sentences.
Subsequently, the posterior model utilizes these relations to generate a target response that incorporates persona sentences without relying on learnable parameters specifically designed to capture NLI relations.

\begin{figure}[t!]
    \centering
    \includegraphics[width=\linewidth]{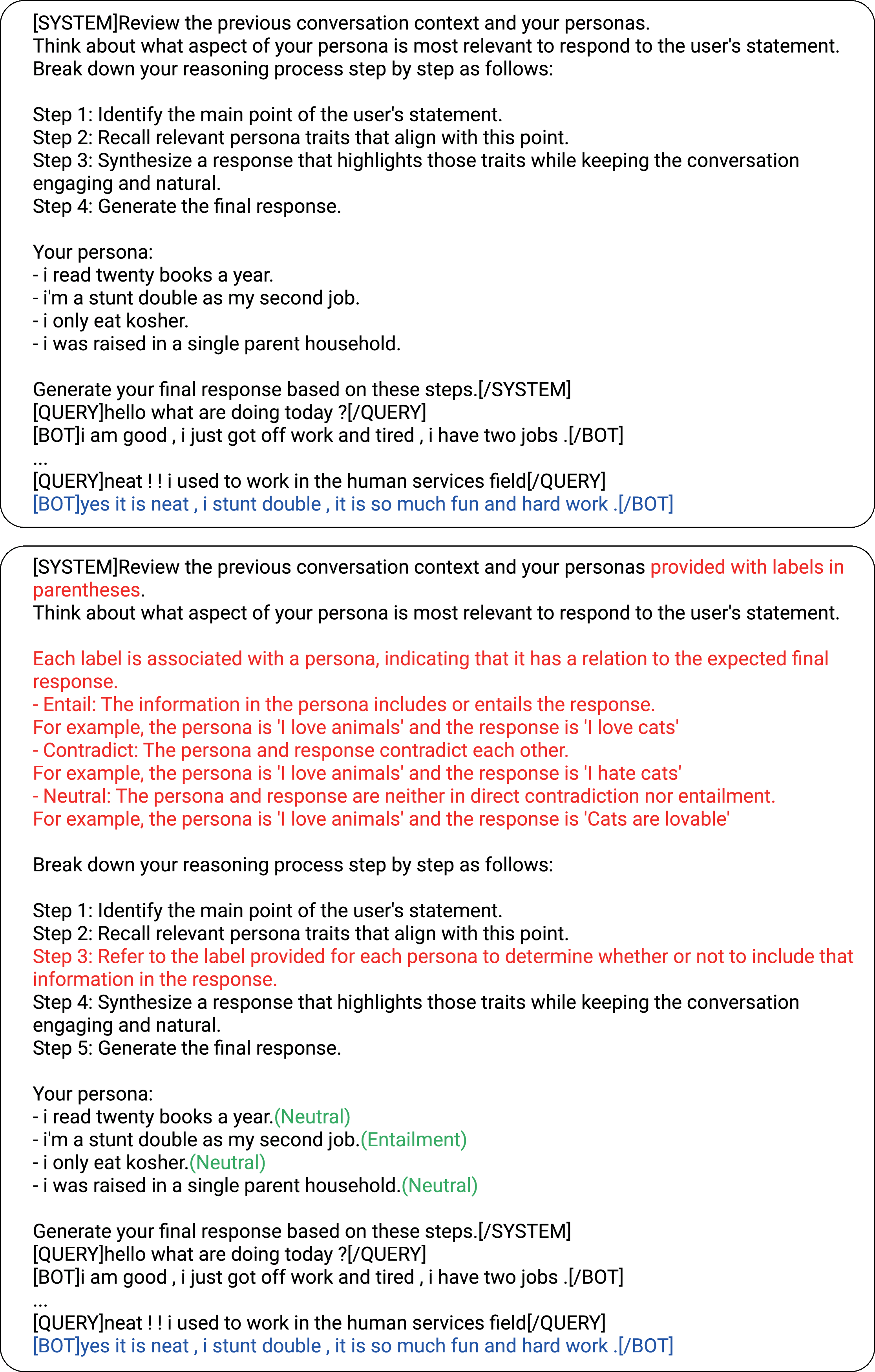}
    \caption{Input prompts for the prior (top) and posterior (bottom) model in MoCoRP LLM.
    [SYSTEM], [QUERY], and [BOT] indicate system message, user query, and bot utterance, respectively.
    A "/" preceding these tokens refers to the end of the corresponding role.
    The texts in red represents the reasoning process that the posterior model needs to additionally reference, while the green text indicates the NLI relations between the persona sentences and the prior response.
    Some utterances of the dialogue history are omitted, and the target response is highlighted in blue.}
    \label{fig:mocorpllm_prompt}
\end{figure}

Inspired by Ouyang et al.~\cite{NEURIPS2022_rlhf}, the LLMs are trained on supervised fine-tuning (SFT) followed by preference learning.
The goal of SFT is to generate a target response $R$ conditioned on persona sentences $P$ and dialogue history $H$, similar to a dialogue setting.
Preference learning is conducted through Direct Preference Optimization (DPO, \cite{NEURIPS2023_dpo}), which requires neither human annotators nor a reward model.
DPO learns a policy to increase the reward for the target response $R$ and decrease the reward for a distractor response $D$, given $P$ and $H$.
To train the MoCoRP LLM, the prior model is trained using SFT and DPO with the given input.
Once training is complete, the prior model generates a prior response, and the NLI expert computes the NLI labels $Z=\{Z^1, Z^2, \dots, Z^k \}$ between the prior response and the persona sentences.
Subsequently, the posterior model additionally takes the NLI labels $Z$ as part of its input, following the same training process as the prior model, using the identical pre-trained LLM backbone.
Both the prior and posterior model input prompts follow a Chain-of-Thought style~\cite{NEURIPS2022_cot}, incorporating step-by-step reasoning, as shown in Figure~\ref{fig:mocorpllm_prompt}.
Due to resource constraints, MoCoRP LLM was trained using LoRA~\cite{hu2022lora}.

\section{Experiments} \label{exp}
\subsection{Dataset} \label{exp:data}
To validate the effectiveness of our model, we conducted experiments using two publicly available persona-based dialogue datasets.
ConvAI2~\cite{dinan-etal-2020-convai} is a multi-turn conversation dataset based on PersonaChat~\cite{zhang-etal-2018-personalizing}, which was collected through crowd-sourcing.
In this dataset, each speaker is required to respond based on the given persona sentences.
MPChat~\cite{ahn-etal-2023-mpchat} is an episodic-memory-based dataset collected from Reddit, designed as a multimodal conversation dataset where persona sentences and dialogue histories are provided alongside images.
This dataset was constructed using rule-based lexical methods and model-based semantic methods and was initially designed for retrieval-based tasks.
However, in this study, we extended its experiments to include generation tasks and report results using text-only information.

Additionally, the Dialogue NLI dataset~\cite{welleck-etal-2019-dialogue} was employed for training the NLI expert and conducting relation learning in the MoCoRP.
This dataset, derived from PersonaChat, is a natural language inference dataset consisting of sentence pairs labeled as entailment, neutral, or contradiction.
For relation learning experiments, we sampled 13K and 1.1K sentence pairs from the ConvAI2 and MPChat datasets, respectively, for training purposes.

\subsection{Method for Comparison} \label{exp:met_comp}
In the experiments, we compared the performance of the proposed model with various baselines across multiple aspects.
For the ConvAI2 dataset, KV Profile Memory and Seq2Seq with attention mechanism are used as official baselines in the ConvAI2 competition~\cite{dinan-etal-2020-convai}.
Lost in Conversation~\cite{golovanov-etal-2019-large} achieves the highest performance in human evaluation during competition, while TransferTransfo~\cite{wolf2019transfertransfotransferlearningapproach} demonstrates the best performance in automatic metrics.
P$^2$BOT~\cite{liu-etal-2020-impress} models dialogue understanding using mutual persona perception between two speakers, and LMEDR~\cite{chen-etal-2023-learning} incorporates external memory to handle entailment and discourse relations, achieving state-of-the-art performance.
For large language models, we utilized Qwen2~\cite{yang2024qwen2technicalreport}, Mistral v0.1~\cite{jiang2023mistral7b}, and instruction-tuned LLaMA 3~\cite{dubey2024llama3herdmodels}, with model sizes of 0.5B, 7B, and 8B, respectively.
For the MPChat dataset, which is a multimodal conversation dataset containing both images and text, we conducted comparisons with baselines that utilize text-only input.
As retrieval-based baselines, TF-IDF score is used and SBERT~\cite{reimers-gurevych-2019-sentence} is utilized as a model designed specifically for semantic search.
Additionally, we report the performance of the BART-large~\cite{lewis-etal-2020-bart}, evaluated in our experimental environment across both datasets.
See Appendix~\ref{app:imp_det} for more detailed implementation settings.

\subsection{Evaluation Metrics} \label{exp:eval_met}
To evaluate the performance of various models, we adopted the following automatic metrics.
Hits@1, Perplexity, and F1 are the official evaluation metrics for the ConvAI2 dataset.
Hits@1 (referred to as R@1 for MPChat) measures the probability that the model assigns the highest score to the correct next dialogue utterance among candidate responses.
Perplexity (PPL) is calculated as the negative log-likelihood of the target response from the given input context, where a lower perplexity indicates a better language model.
F1 measures the word-level precision and recall between the generated response and the target response.

Additionally, we reported BLEU-1 and BLEU-2~\cite{papineni-etal-2002-bleu} to evaluate unigram and bigram overlap, respectively, as well as ROUGE-1 and ROUGE-L~\cite{lin-2004-rouge} to assess unigram overlap and the longest common subsequence between the generated responses and the target responses.
To assess persona consistency, we adopted the C score~\cite{madotto-etal-2019-personalizing}, which utilizes a referee model (in this study, the NLI expert) to evaluate the relations between the generated response $\hat{r}$ and persona $p^i$.
Specifically, the NLI expert categorizes the relation as follows:
\begin{gather}
    \text{NLI}(\hat{r}, p^i) =
    \begin{cases} 
        -1, & \text{if } \hat{r} \text{ contradicts } p^i, \\
        0, & \text{if } \hat{r} \text{ is irrelevant to } p^i, \\
        1, & \text{if } \hat{r} \text{ entails } p^i,
    \end{cases} \\
    \text{C}(\hat{r}) = \sum_{i=1}^k \text{NLI}(\hat{r}, p^i).
    \label{eq15}
\end{gather}
Furthermore, for the MPChat dataset, we also reported Mean Reciprocal Rank (MRR), which calculates the average of the reciprocal ranks of the target response among the candidate responses.

\section{Result} \label{res}
\subsection{Conventional Metric Evaluation} \label{res:con_eval}
\begin{table}
    \caption{Automatic evaluation results on the ConvAI2 dataset.
    The best results are highlighted in bold.
    RL LM [E] and [N, E] indicate cases where the LM loss is applied in relation learning when the relation between premise and hypothesis is entailment, or neutral and entailment, respectively.
    In contrast, RL LM [none] represents cases where the LM loss is not applied during relation learning.}
    \label{tab:convai_lm}
    
    \centering
    \resizebox{\columnwidth}{!}{\begin{tabular}{lcccc}
        \toprule
        Model & Hits@1 ↑ & PPL ↓ & F1 ↑ & C ↑ \\
        \midrule
        
        KV Profile Memory & 55.1 & - & 11.72 & - \\
        Seq2Seq + Attention & 12.5 & 35.07 & 16.82 & - \\
        Lost In Conversation & 17.3 & - & 17.79 & - \\
        TransferTransfo & 82.1 & 17.51 & 19.09 & 12.59 \\
        P$^2$BOT & 81.9 & 15.12 & 19.77 & - \\
        LMEDR & 89.5 & 10.99 & 21.99 & 13.54 \\
        BART (our implementation) & 90.45 & 10.34 & 22.39 & 15.04 \\
        \midrule
        
        MoCoRP (RL LM {[}E{]}) & 90.51 & 10.33 & 22.33 & \textbf{16.06} \\
        MoCoRP (RL LM {[}N,E{]}) & 90.63 & 10.36 & 22.18 & 15.98 \\
        MoCoRP (RL LM {[}none{]}) & \textbf{90.68} & \textbf{10.32} & \textbf{22.49} & 15.96 \\
        \bottomrule
        
        \end{tabular}
    }
\end{table}

\begin{table}
    \caption{Automatic evaluation results on the MPChat dataset.
    The best results are highlighted in bold.
    RL LM [E] and [N, E] indicate cases where the LM loss is applied in relation learning when the relation between premise and hypothesis is entailment, or neutral and entailment, respectively.
    In contrast, RL LM [none] represents cases where the LM loss is not applied during relation learning.}
    \label{tab:mpchat_lm}
    
    \centering
    \resizebox{\columnwidth}{!}{\begin{tabular}{lccccc}
        \toprule
        Model & R@1 ↑ & MRR ↑ & PPL ↓ & F1 ↑ & C ↑ \\
        \midrule
        
        TF-IDF & 10.69 & 18.06 & - & - & - \\
        SBERT & 51.32 & 64.76 & - & - & - \\
        BART (our implementation) & 68.93 & 78.75 & \textbf{28.72} & 14.49 & 33.12 \\
        \midrule
        
        MoCoRP (RL LM {[}E{]}) & \textbf{70.23} & \textbf{79.22} & 29.09 & 14.38 & 33.3 \\
        MoCoRP (RL LM {[}N,E{]}) & 69.49 & 78.87 & 29.13 & \textbf{14.72} & 33.26 \\
        MoCoRP (RL LM {[}none{]}) & 70.19 & 79.13 & 29.16 & 14.16 & \textbf{33.41} \\
        \bottomrule
        
        \end{tabular}
    }
\end{table}

\begin{table*}
    \caption{Comparison of MoCoRP LLM results on the ConvAI2 dataset.
    The suffix 'it' indicates instruction-tuned models.
    Results where the posterior outperforms the prior are underlined, and cases where SFT + DPO achieves better performance than SFT alone are marked with *.}
    \label{tab:convai_llm}
    
    \centering
    \resizebox{\linewidth}{!}{\begin{tabular}{lcccccc}
        \toprule
        Model & F1 ↑ & BLEU-1 ↑ & BLEU-2 ↑ & ROUGE-1 ↑ & ROUGE-L ↑ & C ↑ \\
        \midrule
        
        Qwen2 (SFT, prior) & 20.22 & 16.68 & 6.57 & 20.39 & 18.87 & 18.83 \\
        MoCoRP LLM$_{\text{Qwen2}}$ (SFT, posterior) & \underline{20.23} & \underline{16.71} & \underline{6.67} & \underline{20.42} & \underline{18.91} & \underline{18.95} \\
        Qwen2 (SFT+DPO, prior) & 21.01* & 18.16* & 7.21* & 22.73* & 20.86* & 21.20* \\
        MoCoRP LLM$_{\text{Qwen2}}$ (SFT+DPO, posterior) & \underline{21.09}* & 18.15* & \underline{7.30}* & 22.68* & 20.76* & 20.70* \\
        \midrule
    
        Mistral v0.1 (SFT, prior) & 22.47 & 18.74 & 8.14 & 22.75 & 21.00 & 17.36 \\
        MoCoRP LLM$_{\text{Mistral v0.1}}$ (SFT, posterior) & \underline{22.51} & 18.74 & 8.13 & \underline{22.82} & \underline{21.07} & \underline{17.47} \\
        Mistral v0.1 (SFT+DPO, prior) & 22.84* & 19.24* & 8.42* & 23.65* & 21.74* & 18.57* \\
        MoCoRP LLM$_{\text{Mistral v0.1}}$ (SFT+DPO, posterior) & \underline{22.89}* & \underline{19.27}* & \underline{8.44}* & 23.65* & \underline{21.77}* & \underline{18.62}* \\
        \midrule
    
        LLaMA 3 it (SFT, prior) & 20.91 & 17.40 & 6.92 & 21.52 & 19.81 & 16.46 \\
        MoCoRP LLM$_\text{LLaMA 3 it}$ (SFT, posterior) & \underline{21.03} & \underline{17.54} & \underline{7.03} & \underline{21.74} & \underline{19.92} & \underline{16.57} \\
        LLaMA 3 it (SFT+DPO, prior) & 21.18* & 17.90* & 7.30* & 22.22* & 20.38* & 17.35* \\
        MoCoRP LLM$_\text{LLaMA 3 it}$ (SFT+DPO, posterior) & \underline{21.25}* & \underline{18.05}* & 7.27* & \underline{22.55}* & \underline{20.56}* & \underline{17.78}* \\
        \bottomrule
        
        \end{tabular}
    }
\end{table*}

The experimental results for the ConvAI2 dataset are presented in Table~\ref{tab:convai_lm}, and the results for the MPChat dataset are shown in Table~\ref{tab:mpchat_lm}.
Our method demonstrates overall performance improvements compared to existing baselines, including the previous state-of-the-art.
Notably, there is a significant improvement in the C score, indicating that our approach effectively incorporates persona information into the response during inference, even in the absence of explicit relations, thereby generating a persona-consistent response.
Moreover, consistent performance gains are observed in retrieval-based metrics such as Hits@1, R@1 and MRR.
The results underscore that predicting NLI relations from given persona sentences enhances the ability to select the correct next dialogue utterance among multiple candidates.

We additionally report the results of training the language modeling loss (LM loss) during relation learning (Equation~\ref{eq7}) under various settings.
The LM losses are defined as follows:
RL LM [E] ([E]), where the LM loss is applied to hypotheses that are entailed by the premise;
RL LM [N, E] ([N, E]), where the LM loss is applied to hypotheses that are either entailed or neutral with respect to the premise;
and RL LM [none] ([none]), where the LM loss is not applied during relation learning.

In the ConvAI2 dataset, the model trained with [none] achieves the highest performance in Hits@1, PPL, and F1.
The [N, E] model, however, shows the lowest F1 due to excessive training on the Dialogue NLI dataset, while the [E] model exhibits the worst performance in Hits@1 but achieves the best C score.
Conversely, in the MPChat dataset, the [E] model achieves the best results in R@1 and MRR.
The [N, E] model, which benefits from the LM loss during relation learning, demonstrates the highest F1 but the lowest performance in R@1 and MRR.
The [none] model, on the other hand, achieves the best C score.
Additionally, MoCoRP consistently demonstrates higher PPL compared to BART.

These differences seem to arise from the characteristics and data quantities of the two datasets (see Appendix~\ref{app:data} for details).
The ConvAI2 dataset is relatively data-rich, as each dialogue consists of numerous short utterances.
Therefore, without applying the LM loss during relation learning, the model can achieve optimal performance in this data set.
In contrast, the MPChat dataset comprises fewer but longer utterances per dialogue.
Due to the smaller size of the dialogue dataset, applying LM loss during relation learning helps the model generate responses that exhibit greater textual overlap with the ground truth, resulting in a higher F1.
This highlights that the contribution of different LM losses in relation learning varies depending on the characteristics of the dataset.

Table~\ref{tab:convai_llm} compares the results of MoCoRP LLM using various LLMs on the ConvAI2 dataset.
The proposed model consistently shows that the posterior performance surpasses the prior performance under both SFT and SFT + DPO settings.
Except for minor decreases in specific cases, such as the C score of Qwen2 or the BLEU-2 score of LLaMA 3, posterior performance generally exceeds that of prior.
This demonstrates that the posterior effectively reflects the NLI relations between the response generated by the prior LLM and the persona sentences.
Notably, in all results, MoCoRP LLM consistently exhibits overall performance improvements when both SFT and DPO are used together compared to SFT alone, including the C score.
This suggests that DPO-based preference learning contributes to enhancing response quality by enabling the model to generate more human-like responses.

\subsection{Qualitative Evaluation} \label{res:qual_eval}
Similarity-based metrics, such as ROUGE and BLEU, exhibit relatively low correlations with human judgment and fail to adequately capture aspects such as fluency and consistency~\cite{zhong-etal-2022-towards}.
In contrast, human evaluation provides high accuracy, but incurs significant costs.
To address this issue, we utilized the LLM-based evaluation tool G-Eval~\cite{liu-etal-2023-g} to assess the quality of dialogue responses generated by each model.
For this assessment, we employed OpenAI's \texttt{gpt-4o-2024-08-06} as the evaluation model.
Specifically, 100 responses were sampled from the ConvAI2 dataset for evaluation.
Each response was evaluated 20 times with the given persona sentences and dialogue history.
These measurements were aggregated into scores and weighted averages to calculate overall performance.
Since the official code repository for G-Eval does not provide prompts defined for dialogue response evaluation, we referred to the implementation described in Chiang and Lee~\cite{chiang-lee-2023-closer}.
The evaluation consists of four criteria: Coherence (Coh.), Engagingness (Eng.), Groundedness (Gro.), and Naturalness (Nat.).
Coherence evaluates how well the response maintains the flow of the conversation, with scores ranging from 1 (ignoring context or deviating from the topic) to 3 (fully adhering to the context and maintaining the topic).
Engagingness assesses the level of interest in the response, with scores ranging from 1 (ordinary and dull) to 3 (highly engaging or providing interesting insights).
Groundedness measures how well the response utilizes the given persona sentences, with scores ranging from 0 (no mention of persona) to 1 (effective use of persona).
Naturalness evaluates the naturalness of the response, with scores ranging from 1 (highly unnatural) to 3 (highly natural).

As shown in Table~\ref{tab:convai_geval} and Table~\ref{tab:mpchat_geval}, the proposed model demonstrates consistent improvements across all evaluation criteria in both ConvAI2 and MPChat.
Notably, the proposed model exhibits a significant improvement in Engagingness for ConvAI2 dataset compared to the existing baselines.
In the MPChat dataset, which features a more conversational and informal tone, the model demonstrates substantial enhancements in Naturalness.
Moreover, Groundedness and Naturalness surpass the ground-truth results in both datasets.
These results indicate that MoCoRP not only effectively conditions on the persona sentences to generate natural responses, but also maintains engaging and conversational flows.
These findings indicate that MoCoRP effectively conditions on persona sentences to generate natural and engaging responses while maintaining conversational flows.

\begin{table}
    \caption{Qualitative evaluation results on the ConvAI2 dataset.
    The best results are highlighted in bold.
    Cases where the proposed model outperforms the ground-truth results are marked with *.}
    \label{tab:convai_geval}
    
    \centering
    \resizebox{\columnwidth}{!}{\begin{tabular}{lcccccc}
        \toprule
        Model & Coh. ↑ & Eng. ↑ & Gro. ↑ & Nat. ↑ & Overall ↑ \\
        \midrule
    
        TransferTransfo & 1.83 & 1.47 & 0.33 & 1.83 & 1.36 \\
        LMEDR & 2.07 & 1.67 & 0.41 & 2.22 & 1.59 \\
        BART (our implementation) & 2.13 & 1.74 & 0.47 & 2.35 & 1.67 \\
        MoCoRP & \textbf{2.16} & \textbf{1.85} & \textbf{0.49}* & \textbf{2.39}* & \textbf{1.72} \\
        \midrule
        
        Ground-truth & 2.27 & 1.93 & 0.46 & 2.33 & 1.75 \\
        \bottomrule
        
        \end{tabular}
    }
\end{table}

\begin{table}
    \caption{Qualitative evaluation results on the MPChat dataset.
    The best results are highlighted in bold.
    Cases where the proposed model outperforms the ground-truth results are marked with *.}
    \label{tab:mpchat_geval}
    
    \centering
    \resizebox{\columnwidth}{!}{\begin{tabular}{lcccccc}
        \toprule
        Model & Coh. ↑ & Eng. ↑ & Gro. ↑ & Nat. ↑ & Overall ↑ \\
        \midrule
    
        BART (our implementation) & 1.93 & 1.64 & 0.65 & 2.16 & 1.60 \\
        MoCoRP & \textbf{2.08} & \textbf{1.68} & \textbf{0.68}* & \textbf{2.25}* & \textbf{1.67} \\
        \midrule
        
        Ground-truth & 2.13 & 1.73 & 0.62 & 2.21 & 1.67 \\
        \bottomrule
        
        \end{tabular}
    }
\end{table}

\subsection{Ablation Study} \label{res:abl}
To investigate the impact of relation learning on performance improvement, we gradually increased the proportion of relation learning training data.
For ConvAI2, the proportion ranged from 0\% to 60\%, while for MPChat, it ranged from 0\% to 30\%.
The changes in performance are presented in Table~\ref{tab:convai_abl}.
Notably, in response selection metrics such as Hits@1, R@1, and MRR, the proposed model consistently demonstrates high performance across both datasets.
In the ConvAI2 dataset, the proposed MoCoRP records the lowest PPL, indicating efficient language modeling performance, while achieving performance close to the best in F1 and C score.
No clear trend is observed between the amount of relation learning data and performance, suggesting that the dataset is relatively less sensitive to the amount of data used for relation learning.
In contrast, the MPChat dataset shows that the proposed model exhibits relatively higher PPL than others, but achieves the highest performance in C score.
When relation learning is not applied (RL 0\%), the model exhibits overall lower performance, achieving the highest performance at 5\%, followed by a gradual decline as the amount of training data increases.
This suggests that MPChat is relatively more sensitive to the amount of relation learning data.
These differences in performance are attributed to the different characteristics and data quantities of the two datasets, as discussed earlier.
Due to its abundant dialogue data, the ConvAI2 dataset remains relatively insensitive to the amount of relation learning data.
On the other hand, the MPChat dataset experiences a significant performance decline when extensive relation learning is applied, owing to its limited dialogue data.
This suggests that performance varies depending on the characteristics of each dialogue dataset, and the amount of data used for relation learning should be appropriately selected.
Therefore, using relation learning training data equivalent to 10\% of the dialogue dataset for ConvAI2 and 5\% for MPChat yields the best performance for our experiments.

\begin{table}
    \caption{Ablation study on ConvAI2 dataset.
    The best results are highlighted in bold.
    RL x\% denotes the proportion of relation learning training data as a percentage of the corresponding persona-based dialogue dataset.
    The proposed MoCoRP is indicated with *.}
    \label{tab:convai_abl}
    
    \centering
    \resizebox{\columnwidth}{!}{\begin{tabular}{llccccc}
        \toprule
        Dataset & Model & Hits@1 (R@1) ↑ & MRR ↑ & PPL ↓ & F1 ↑ & C ↑ \\
        \midrule
        
        \multirow{5}{*}{ConvAI2} & MoCoRP (RL 0\%) & 90.46 & - & 10.36 & 22.53 & 16 \\
        & MoCoRP (RL 20\%) & 90.36 & - & 10.37 & 22.32 & 15.86 \\
        & MoCoRP (RL 40\%) & 90.64 & - & 10.36 & \textbf{22.56} & 15.91 \\
        & MoCoRP (RL 60\%) & 90.58 & - & 10.37 & 22.35 & \textbf{16.02} \\
        \cmidrule{2-7}
        
        & MoCoRP (RL 10\%)* & \textbf{90.68} & - & \textbf{10.32} & 22.49 & 15.96 \\
        \midrule
        
        \multirow{5}{*}{MPChat} & MoCoRP (RL 0\%) & 67.14 & 77.28 & 28.73 & 14.49 & 31.93 \\
        & MoCoRP (RL 10\%) & 69.53 & 78.99 & \textbf{28.51} & 14.53 & 31.58 \\
        & MoCoRP (RL 20\%) & 68.45 & 78.15 & 28.9 & \textbf{14.69} & 33.11 \\
        & MoCoRP (RL 30\%) & 66.8 & 76.85 & 28.61 & 14.3 & 31.91 \\
        \cmidrule{2-7}
        
        & MoCoRP (RL 5\%)* & \textbf{70.23} & \textbf{79.22} & 29.09 & 14.38 & \textbf{33.3} \\
        \bottomrule
        
        \end{tabular}
    }
\end{table}

\subsection{Case Study} \label{res:case_std}
\begin{table}
    \caption{Case study on ConvAI2 dataset.
    Q and R represent the user query and the previous bot utterance, respectively.}
    \label{tab:case_std_convai2:1}
    
    \centering
    \resizebox{\columnwidth}{!}{\begin{tabular}{ll}
        \toprule
        \multirow{4}{*}{Persona} & i volunteer at a soup kitchen. \\
        & cheeseburgers are my favorite food. \\
        & i was poor growing up. \\
        & i like watching war documentaries. \\
        \midrule
        
        \multirow{9}{*}{Context} & Q: hello , how are you tonight ? do you have pink and blue hair ? \\
        & R: hi , i'm doing pretty good these evening \\
        & Q: what do you like to do in your spare time ? i bird watch . \\
        & R: i actually have purple hair and i love it and enjoy watching war documentaries \\
        & Q: those are fun . i've a cat , do you ? \\
        & R: i've a bird and she loves cheeseburgers like me , my favorite \\
        & Q: is that healthy for birds ? \\
        & R: probably not but a bite here and there i think is okay \\
        & Q: do you like the holidays ? i don't . \\
        \midrule
        
        TransferTransfo & i love the holidays but i don ' t like the cold \\
        LMEDR & i do not like the holidays either , i was poor growing up so i never celebrated \\
        BART & i do not but i do like volunteering at the soup kitchen \\
        MoCoRP & yes i do, i love the holidays. i volunteer at a soup kitchen during them \\
        Ground-truth & i do love the holidays i volunteer a soup kitchen during the holiday season \\
        \bottomrule
        
        \end{tabular}
    }
\end{table}

We conducted a case study to compare the responses generated by models trained using different methodologies.
Tables~\ref{tab:case_std_convai2:1}, \ref{tab:case_std_convai2:2}, and \ref{tab:case_std_mpchat} present the results for the persona-based dialogue datasets.
Across both datasets, our approach demonstrates the ability to generate contextually appropriate responses while effectively utilizing the given persona.

As illustrated in Table~\ref{tab:case_std_convai2:1}, for the ConvAI2 dataset, TransferTransfo repeats phrases from the query, producing contextually irrelevant sentences.
Although both LMEDR and BART successfully integrate personas like "I was poor growing up" and "I volunteer at a soup kitchen", their generated responses conflict with the ground-truth responses.
In comparison, our model not only incorporates the persona effectively, but also produces contextually relevant responses that closely match the ground-truth.

In Table~\ref{tab:case_std_convai2:2}, examples demonstrate that the proposed model does not excessively rely on persona information.
Specifically, TransferTransfo fails to comprehend the context and repeatedly responds based on persona information.
Similarly, LMEDR echoes both user queries and persona in its replies.
BART tends to repeat previous bot utterances, indicating a lack of contextual understanding.
In contrast, the proposed MoCoRP aligns closely with the ground-truth results by effectively understanding the context, avoiding over-reliance on persona, and guiding the conversation toward different topics.

Table~\ref{tab:case_std_mpchat} shows examples from the MPChat dataset, which includes more colloquial language and abbreviations.
In this case, BART fails to fully comprehend the context, generating a response that is related to the persona but lacks contextual relevance.
On the other hand, our model leverages both the persona and the context to generate a coherent and contextually suitable response.

\begin{table}
    \caption{Case study on ConvAI2 dataset.
    Q and R represent the user query and the previous bot utterance, respectively.}
    \label{tab:case_std_convai2:2}
    
    \centering
    \resizebox{\columnwidth}{!}{\begin{tabular}{ll}
        \toprule
        \multirow{5}{*}{Persona} & i am 60years old. \\
        & i think i'll retire in a few years. \\
        & i'm a librarian. \\
        & i've visited spain a times. \\
        & i really like to travel. \\
        \midrule
        
        \multirow{9}{*}{Context} & Q: hello , how are you today ? \\
        & R: i am very well thank you . how are you today ? \\
        & Q: i'm great . i just got back from a nice long walk . \\
        & R: i'm thinking of retiring soon so i will be able to walk more . \\
        & Q: it is a fantastic way to clear your mind and get exercise at the same time . \\
        & R: do you like to travel ? i have been to spain several times . \\
        & Q: i do like to travel , though i've not been to spain . \\
        & R: it is hard to do on a librarians salary sometimes . \\
        & Q: i can understand the struggle \\
        \midrule
        
        TransferTransfo & i have been to spain and i like it . \\
        LMEDR & i understand . i'm 60 so it is hard for me to travel now . \\
        BART & it is hard to do on a librarians salary sometimes. \\
        MoCoRP & what do you do for fun? \\
        Ground-truth & do you have any hobbies or a job ? \\
        \bottomrule
        
        \end{tabular}
    }
\end{table}

\begin{table}
    \caption{Case study on MPChat dataset.
    Q and R represent the user query and the previous bot utterance, respectively.
    \textit{itap} is an abbreviation for "I Took A Picture" and \textit{BTW} stands for "By The Way".}
    \label{tab:case_std_mpchat}
    
    \centering
    \resizebox{\columnwidth}{!}{\begin{tabular}{ll}
        \toprule
        \multirow{4}{*}{Persona} & my hibiscus trees are blooming \\
        & my beautiful cat hanging out in the yard \\
        & loving my current combo - sony a6000 with sigma 16mm f1.4 lens. \\
        & one of my all time favorite sunset shots, taken on my property in southern michigan. \\
        \midrule
        
        \multirow{2}{*}{Context} & R: itap of flowers at sunrise \\
        & Q: what camera? stunning picture btw \\
        \midrule
        
        BART & thank you! it was taken on my property in southern michigan. \\
        MoCoRP & thanks! sony a6000 with sigma 16mm f1.4 lens \\
        Ground-truth & thank you. sony a6000 and sigma 16mm lens. \\
        \bottomrule
        
        \end{tabular}
    }
\end{table}

\section{Conclusion} \label{con}
In this study, we propose a framework that produces explicit relations between persona sentences and responses and integrates them into language models. By incorporating NLI relations, predicted by a dedicated classifier, into dialogue agents, our approach enhances persona consistency and enables contextually engaging and coherent dialogue generation. We apply this framework to a pre-trained language model like BART, and furthermore, extend it to modern large language models through alignment tuning. Experiments on two public datasets demonstrate that our model achieves superior performance, particularly in metrics related to retrieval accuracy and generation coherence.

\backmatter

\noindent

\begin{appendices}

\section{Statistics of Persona-based Dialogue Dataset} \label{app:data}

Table~\ref{tab:stat_data} presents the statistics for two persona-based dialogue datasets.
ConvAI2 consists of approximately 19,000 crowd-sourced dialogues, each containing approximately 15 utterances, resulting in a total of approximately 140,000 examples.
Among these, approximately 130,000 examples are included in the training dataset.
MPChat, in contrast, is an episodic-memory-based dataset collected from Reddit, containing 15,000 dialogues.
Each dialogue includes approximately 3 utterances, resulting in a total of about 23,500 examples, with around 19,000 used for training.
While ConvAI2 is abundant in examples and features dialogues with many short utterances and persona sentences, MPChat has fewer examples and dialogues, each comprising fewer but longer persona sentences and utterances.

\begin{table}
    \caption{Statistics of persona-based dialogue dataset.
    The numbers in parentheses indicate the size of the training set.
    The avg. persona tokens refers to the average token length of each persona sentence, and the avg. utterance tokens refers to the average token length of each utterance.
    The token counts were measured using the BART tokenizer.}
    \label{tab:stat_data}
    
    \centering
    \resizebox{\columnwidth}{!}{\begin{tabular}{lccccc}
        \toprule
        Dataset & \# dialogues & \# utterance & \# examples & avg. persona tokens & avg. utterance tokens \\
        \midrule
        
        ConvAI2 & 18,878 & 278,478 & 139,239 (131,438) & 7.68 & 11.92 \\
        MPChat & 15,000 & 42,531 & 23,501 (18,904) & 11.58 & 20.72 \\
        \bottomrule
        
    \end{tabular}
    }
\end{table}

\section{Statistics of Relations} \label{app:stat_rel}

\begin{table}
    \caption{Statistics of relations in the training set of each dataset.
    For ConvAI2 and MPChat, the NLI expert predicts the relations between persona sentences and target responses.
    In contrast, Dialogue NLI represents the NLI labels assigned to premises and hypotheses.}
    \label{tab:stat_rel}
    
    \centering
    \resizebox{\columnwidth}{!}{\begin{tabular}{lccc}
        \toprule
        Dataset & \# Entailment & \# Neutral & \# Contradiction \\
        \midrule
        
        ConvAI2 & 83,153 (14.08\%) & 496,710 (84.08\%) & 10,882 (1.84\%) \\
        MPChat & 20,488 (26.66\%) & 55,204 (71.85\%) & 1,143 (1.49\%) \\
        \midrule
        
        Dialogue NLI & 100,000 (32.25\%) & 100,000 (32.25\%) & 110,110 (35.50\%) \\
        \bottomrule
        
    \end{tabular}
    }
\end{table}

Table~\ref{tab:stat_rel} illustrates the proportions of NLI relations (entailment, neutral, contradiction) in the training set of each dataset.
In both ConvAI2 and MPChat, neutral relations dominate, accounting for approximately 85\% and 70\% of the data, respectively.
Entailment relations are less frequent, representing around 15\% in ConvAI2 and 25\% in MPChat.
Contradictions are extremely rare, making up only about 1.5\% in both datasets.
The Dialogue NLI dataset, used for relation learning, exhibits a balanced distribution of relations, with each type of relation comprising roughly 30\%.
This balanced distribution enables pre-training on Dialogue NLI to provide dialogue models with a more diverse and comprehensive representation of NLI relations.

\section{Implementation Details} \label{app:imp_det}
To address the context length limitations of the models and ensure efficient training, the maximum turn length of the dialogue history was limited to 14.
For the ConvAI2 dataset, the number of distractor responses $l$ was 19, while for the MPChat dataset, it was 99.
During training, three distractor responses were randomly selected along with the target response and provided as input to the models.
In contrast, during the evaluation, all distractors were used to assess the performance.

The NLI expert was trained on the Dialogue NLI dataset with a batch size of 64 and a learning rate of 1e-5 for one epoch, achieving an accuracy of 92.43\% on the test set.
The BART model was fine-tuned on the ConvAI2 and MPChat datasets during both relation learning and dialogue learning.
For these datasets, batch sizes were set to 4 and 2, respectively, with learning rates of 8e-6 and 1e-5, and trained for 3 and 5 epochs, respectively.
Early stopping was applied for the MPChat dataset, halting training at the completion of the third epoch.
For relation learning, 10\% of the fine-tuning dataset was used for ConvAI2, corresponding to approximately 13,000 samples, while 5\% was used for MPChat, corresponding to approximately 1,100 samples.
The coefficient $\alpha$ for relation prediction was set to 0.1 during training.

All LLMs were trained with LoRA applied to all linear layers, using a rank of 8 and an alpha of 16.
The Qwen2 0.5B model was fine-tuned with a batch size of 128, a learning rate of 5e-5, and trained for 5 epochs during SFT, while a batch size of 64 was used for DPO.
The Mistral v0.1 7B model was fine-tuned with a batch size of 128, a learning rate of 1e-5, and trained for 3 epochs during SFT, with a batch size of 64 for DPO.
The instruction-tuned LLaMA3 8B was fine-tuned with a batch size of 64, a learning rate of 1e-5, and trained for 3 epochs during SFT, with a batch size of 32 for DPO.
During DPO training, all models were trained for one epoch with a learning rate set to 0.01 times the SFT learning rate.
To ensure fairness, the same hyperparameters were applied to both the prior and posterior models.
The posterior model utilized NLI relations derived from the outputs of the prior model after completing the DPO training.

All models were trained using the AdamW optimizer~\cite{loshchilov2019decoupledweightdecayregularization}.
The MoCoRP was trained on a single NVIDIA GeForce RTX 3090 GPU with 24 GB of memory, whereas the MoCoRP LLM was trained using eight NVIDIA A100 Tensor Core GPUs with 40 GB of memory each.
All experiments were implemented and conducted using PyTorch and Hugging Face libraries, including Transformers \cite{wolf-etal-2020-transformers}, Accelerate, TRL, and PEFT.
The evaluation was performed using the ParlAI library \cite{miller-etal-2017-parlai}.
\end{appendices}

\bibliography{sn-bibliography}

\section*{Statements \& Declarations}
\begin{itemize}
\item Author Contributions: All authors contributed to the study conception. Methodology was performed by Kyungro Lee, Dongha Choi, and Hyunju Lee. The first draft of the manuscript was written by Kyungro Lee. Dongha Choi and Hyunju Lee commented on previous versions of the manuscript. All authors have read and approved the final manuscript. Hyunju Lee acquired funding and supervised the study.

\item Funding: This research was supported by a National Research Foundation of Korea (NRF) grant funded by the Korean government (MSIT) (2021R1A2C2006268) and Institute of Information communications Technology Planning Evaluation (IITP) grant funded by the Korea government (MSIT) [No.2019-0-01842, Artificial Intelligence Graduate School Program (GIST)].

\item Competing Interests: The authors declare that they have no known financial or personal conflicts of interest that could have influenced the work reported in this paper.

\item Data Availability: The datasets supporting the findings of this study are publicly available at \url{https://parl.ai/projects/convai2/}, \url{https://github.com/ahnjaewoo/MPCHAT}, and \url{https://wellecks.com/dialogue_nli/}.

\item Code Availability: The code used in this study is publicly available at \url{https://github.com/DMCB-GIST/MoCoRP}.
\end{itemize}

\end{document}